\documentclass[10pt,twocolumn,letterpaper]{article}

\usepackage[pagenumbers]{cvpr}

\usepackage{array}
\usepackage{ctable}
\usepackage[ruled]{algorithm2e}
\usepackage{rotating}
\usepackage{multirow}
\usepackage{booktabs}
\newcommand{\B}{\bf}
\usepackage[pagebackref,breaklinks,colorlinks]{hyperref}

\usepackage{amsmath}

\usepackage[T1]{fontenc}
\usepackage[utf8]{inputenc}
\usepackage[american]{babel}

\usepackage{graphicx}
\usepackage{xcolor}
\graphicspath{{images/}}
\DeclareGraphicsExtensions{.eps}
\DeclareGraphicsExtensions{.jpg}
\DeclareGraphicsExtensions{.pdf}

\usepackage{xspace}

\newcommand*{\helvetica}{\fontfamily{phv}\selectfont\scriptsize}

\newcommand{\one}{({\em i}\/)\xspace}
\newcommand{\two}{({\em ii}\/)\xspace}
\newcommand{\three}{({\em iii}\/)\xspace}
\newcommand{\four}{({\em iv}\/)\xspace}

\newcommand{\SN}[1]{SoccerNet-v{#1}\xspace}
\newcommand{\Sota}{State-of-the-art\xspace}

\definecolor{royalblue}{rgb}{0, 0.443137255, 0.737254902}

\newcommand{\boldVerticalLine}{\color{black}\vrule width 0.75pt}

\usepackage[normalem]{ulem}

\AtBeginDocument{%
  \providecommand\BibTeX{{%
    \normalfont B\kern-0.5em{\scshape i\kern-0.25em b}\kern-0.8em\TeX}}}

\usepackage[capitalize]{cleveref}
\crefname{section}{Sec.}{Secs.}
\Crefname{section}{Section}{Sections}
\Crefname{table}{Table}{Tables}
\crefname{table}{Tab.}{Tabs.}

\def\cvprPaperID{*****} 
\def\confName{CVPR}
\def\confYear{2022}

\begin{document}

\title{A Graph-Based Method for Soccer Action Spotting Using Unsupervised Player Classification}

\author{Alejandro Cartas\\
\and\vspace{5mm}
Coloma Ballester\\
Universitat Pompeu Fabra\\
Barcelona, Spain\\
{\tt\small \{firstname.lastname\}@upf.edu}
\and
Gloria Haro\\
}
\maketitle

\begin{abstract}
Action spotting in soccer videos is the task of identifying the specific time when a certain key action of the game occurs. Lately, it has received a large amount of attention and powerful methods have been introduced. Action spotting involves understanding the dynamics of the game, the complexity of events, and the variation of video sequences. Most approaches have focused on the latter, given that their models exploit the global visual features of the sequences. In this work, we focus on the former by (a) identifying and representing the players, referees, and goalkeepers as nodes in a graph, and by (b) modeling their temporal interactions as sequences of graphs. For the player identification, or player classification task, we obtain an accuracy of $97.72$\% in our annotated benchmark. For the action spotting task, our method obtains an overall performance of $57.83$\% average-mAP by combining it with other audiovisual modalities. This performance surpasses similar graph-based methods and has competitive results with heavy computing methods. Code and data are available at \url{https://github.com/IPCV/soccer_action_spotting}.
\end{abstract}
\section{Introduction}
\label{sec:intro}

Soccer is perhaps the most popular sport in the world as it is officially played in 187 countries~\cite{FPFD2019FIFAReport} and, as an example, it gathered an estimated live audience of 1,12 billion people for the 2018 World Cup final ~\cite{FIFA2018AudienceSummary}, a figure close to fifteen percent of the world population. Moreover, this interest in broadcast soccer matches has become one of the major sources of revenue for soccer clubs in leagues such as the European \cite{KPMG2020EuropeanChampions}. Accordingly, there is a growing interest in analyzing soccer videos to provide tactical information \cite{Anzer2022TacticalPatterns}, quantitative statistics and visualizations \cite{Arbues2020Always}, camera selection in broadcasting \cite{Chen2018CameraSelection}, or video summaries \cite{Giancola2018SoccerNetv1} that aim to highlight the main events of a match. In all these tasks, the automatic spotting of key actions such as goal, corner or red card (among others) stands as a fundamental tool.

At the present time, locating events in soccer videos is manually performed by trained operators aided by algorithmic tools. These operators not only rely on recorded broadcast videos, but also in event metadata coming from logs, GPS data, and video-tracking data \cite{Pappalardo2019PublicData}. This is a laborious task considering that a soccer match lasts around 90 minutes, but it can contain more than 1,500 events \cite{Sanabria2020Profiling} and take about 8 hours to annotate them \cite{Giancola2018SoccerNetv1,Zhou2021FeatureCombination}.

As defined in \cite{Giancola2018SoccerNetv1}, the action spotting task aims at locating the anchor time that identifies such event. Current approaches for action spotting from soccer videos are based on modeling the dynamics of the game, the complexity of the events and the variation of the video sequences. Several methods have been focused on the latter \cite{Sanabria2019Deep,Giancola2018SoccerNetv1,Cioppa2020context,Zhou2021FeatureCombination}, as they exploit the similar visual features of \emph{standard} shots of video sequences. For instance, the shots of a standard \emph{goal scoring} video sequence could consist of a goal-view of a player shooting and scoring, a close-up of fans rejoicing, and a zoom-in of the team celebrating. Other methods have considered also modeling the narration given by the broadcast commentators \cite{Vanderplaetse2020Improved}. In contrast, a few methods \cite{Cioppa2021CameraCalibration} have modeled the visual dynamics of the game as it entails understanding the interactions of the players according to the rules of the game. Moreover, videos of broadcast sport events present an additional difficulty since they usually do not cover all the field, making the location of the players unknown at a given time. Besides, broadcast video sequences are not always showing the main action of the game, as they also capture the audience in the stands, the coaches in the bench, and replays.

\begin{figure*}[!t]
\centering
\includegraphics[height=8cm]{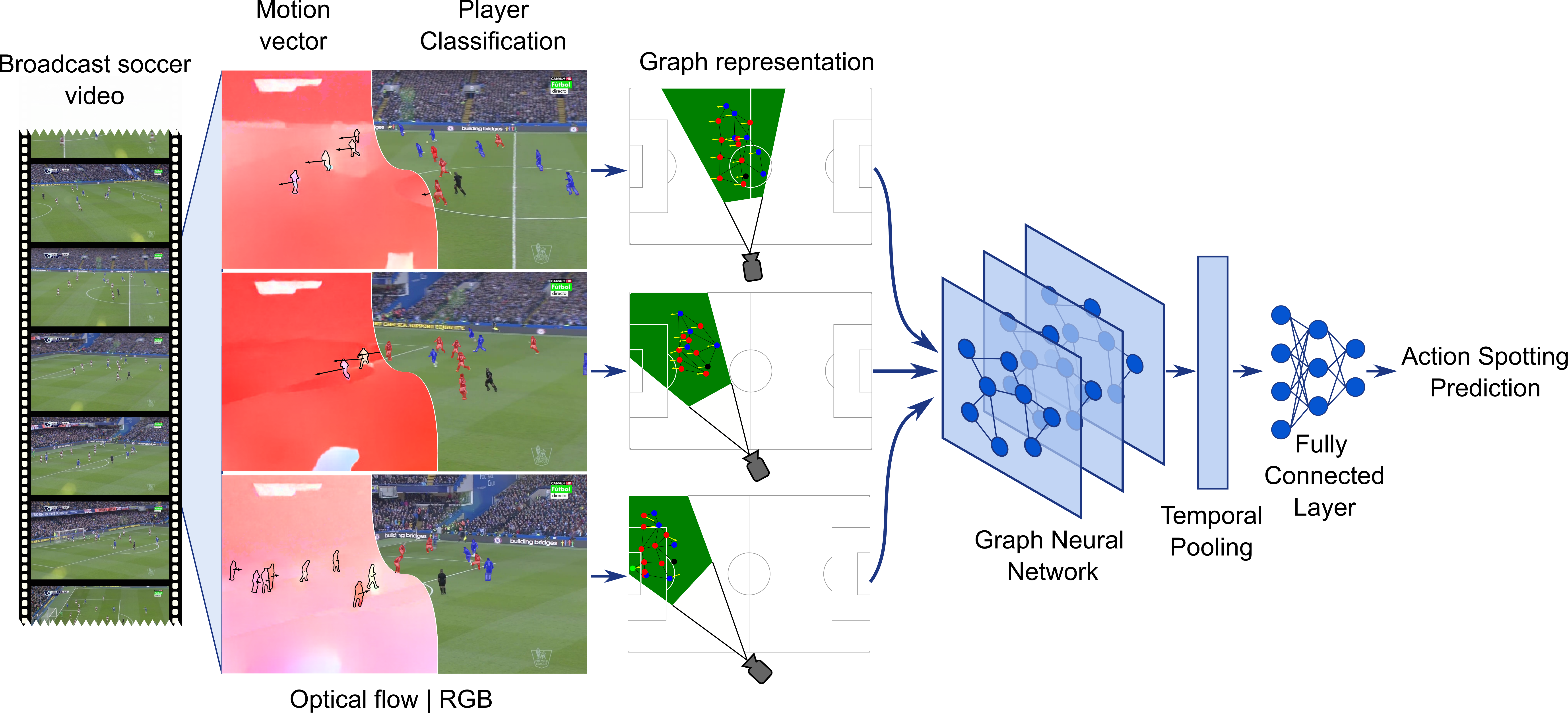}
\caption{Overview of our approach. Given a sequence of frames from a soccer match, for all the people on the field we calculate their motion vector and classify them in \emph{referees}, \emph{players}, and \emph{goalkeepers}. Using this information and their location provided by a camera calibration model, we create a graph representation of the seen scene. These graphs are later processed by our proposed architecture to spot an action.}
\label{fig:overview}
\end{figure*}

In this work, our aim is to model the interactions of soccer players rather than using global generic features to perform action spotting in soccer videos. Consequently, we propose a model that represents the players in the field as nodes linked by their distance in a graph. The information about each player contained in the nodes is their position on the field, a calculated motion vector, and their predicted player class. A schematic overview of our approach is shown in Figure \ref{fig:overview}. 

\paragraph{Contributions} Our contributions are summarized as follows: \one We propose a novel graph-based architecture for action spotting in soccer videos that models the players and their dynamics in the field. The representation of the players relies on their extracted localization, predicted classification, and calculated motion. \two In creating this representation, we also propose an unsupervised method for player classification robust to different sample sizes. This method is based in the segmentation and location of the players on the field. \three We provide an extensive evaluation on the graph representation detailing the contribution of each individual input. \four We show how the results of the graph-based model can be boosted when combined with audiovisual features extracted from the video. Our evaluation shows a better performance than similar methods.

\section{Related Work}
\label{sec:related}

\paragraph{Action spotting and video summarization in sports} The objective of video summarization in sports is to highlight key moments from a game, for instance, when a team scores or when fouls are committed. Relevant work on soccer video summarization includes \cite{Coimbra2018Shape,Sanabria2019Deep}. In \cite{Coimbra2018Shape}, visual highlights of a soccer match are extracted by calculating importance scores of audio and metadata. In \cite{Sanabria2019Deep}, a soccer match is summarized by using a hierarchical recursive neural network over sequences of \emph{event} proposals extracted from frames. In \cite{Giancola2018SoccerNetv1}, the task of sports video summarization is cast as spotting actions in video sequences, a task similar to action detection. Moreover, they introduced SoccerNet, the largest dataset of soccer matches for video summarization. The authors propose a neural network architecture that extracts visual features from sequences of frames and spots actions by using a NetVlad \cite{Arandjelovic16} pooling layer. This architecture was further extended by combining audiovisual modalities \cite{Vanderplaetse2020Improved} and by preserving the temporal context of the pooling layer independently \cite{Giancola2021NetVLADPlusPlus}. Another work on the same dataset was presented in \cite{Cioppa2020context}. They introduce a context-aware loss function along with a deep architecture for video summarization that consists of a feature segmentation and action spotting modules. A closer work to ours was presented in \cite{Cioppa2021CameraCalibration}. They model the players in a graph using their estimated location in the soccer pitch using a camera calibration approach \cite{Sha2020CameraCalibration}. Their architecture fusions the output of a graph neural network \cite{Li2021DeepGCNs} using their previous model \cite{Cioppa2020context}. Besides having different mechanisms for spotting actions, our graph representation is more robust by classifying the players and calculating their motion vector. Finally, \cite{Zhou2021FeatureCombination} presents a transformer based architecture that uses as input features from five action recognition methods. In contrast, our approach models the dynamics of the soccer players rather than using global classification features from the frames and it is not as computationally demanding.

\paragraph{Automatic player classification} A source of information for semantic understanding of sports videos comes from player classification. It has been exploited in other tasks including player detection \cite{Manafifard2016MultiplayerDetection}, player tracking \cite{Tong2011LabelingTracking}, team activity recognition \cite{Bialkowski2013TeamActivities}. Traditional methods for automatic player classification rely on the combination of color histograms (RGB~\cite{Mazzeo2012football}, L$^*$a$^*$b$^*$~\cite{Bialkowski2013TeamActivities}, L$^*$u$^*$v$^*$~\cite{Tong2011LabelingTracking}) and clustering algorithms ($k$-means~\cite{Tran2012LongView}, Gaussian Mixture Models (GMMs)~\cite{Montanes2012RealTime}, Basic Sequential Algorithmic Scheme (BSAS)~\cite{DOrazio2009BSAS}). A more recent supervised approach \cite{Istasse2019Associative} performs a semantic segmentation of players that learns a separation embedding between the teams in the field. In comparison with this method, our method does not rely on pixel-level annotations for training the network. Another recent approach \cite{Koshkina2021contrastive} trains an embedding network for team player discrimination using a triplet loss for contrastive learning. Contrary to this approach, our method works on partial views of the field and not only clusters the two majority classes (\emph{players from team 1} and \emph{team 2}) but also considers minority classes (\emph{referees} and \emph{goalkeepers}).

\paragraph{Graphs for sports analytics} Graphs have been widely used for modeling team ball sports as they represent the players as nodes of a network and their interactions as edges, for example in water polo~\cite{Passos2010Networks}, volleyball~\cite{Qi2020stagNet}, soccer~\cite{Anzer2022TacticalPatterns,Stockl2021Offensive}, etc. Graphs can capture patterns of a game at individual, group, and global level \cite{Baldu2019Using}. As an example of the global level, in \cite{Qi2020stagNet} it is presented a deep architecture that models volleyball players and their individual actions to predict team activities using recursive networks and spatio-temporal attention mechanisms. At the group level, in \cite{Anzer2022TacticalPatterns} it is proposed a graph model that detects tactical patterns from soccer games videos by tracking each player and using variational autoencoders. At the individual level, in \cite{Stockl2021Offensive} defensive quality of individual soccer players is measured by using pitch-player passing graph networks. Since our graph-based model only relies on visual information, our approach involves locating the players in the pitch and classifying them in comparison with \cite{Anzer2022TacticalPatterns}. Moreover, our graph representation only covers the players on the scene as the pitch is not fully visible on broadcast games such as volleyball \cite{Qi2020stagNet}.

\section{Proposed Approach}
\label{sec:method}

Our approach to spot actions from soccer videos relies in representing the seen players on a frame as nodes in a graph. These nodes encode the location of the players, their motion vector, and player classification. In this representation, a player shares an edge with another player according to their approximated physical distance. All this information is extracted by first segmenting the players and locating them in the field using a camera calibration method. In the next step, the five categories of players are automatically classified by finding appropriate field views and using color histograms from their segmentation. The motion vector of a player is calculated as the dominant motion vector from their segmented optical flow mask. Once all the graphs from a sequence of frames are extracted, they are processed by our graph neural network as illustrated in Figure \ref{fig:overview}. All the steps of our approach are thoroughly detailed in the following sections.

\subsection{Players Segmentation and Localization}
\label{sec:playerSegmentationLocalization}

\paragraph{Players Segmentation} Our approach relies on the detection and segmentation of players and referees because their pixel regions allows us to extract useful information and help to disambiguate players in the case of occlusions. Specifically, we leverage player regions to improve their uniforms clustering and estimate the motion of each player from the estimated optical flow in those regions, thus, disregarding the motion of background pixels close to a certain player. We use PointRend~\cite{Kirillov2020PointRend} to perform a semantic segmentation on the soccer videos and filter out all the categories other than \emph{person}.

\paragraph{Players Localization} Our method also relies on the position of the players and referees inside the soccer pitch. Their position is not only used inside our action spotting model but also in the player uniforms clustering, as further explained in Section \ref{sec:playerClassification}. We consider that a player/referee is \emph{touching} the ground at the midpoint between the bottom coordinates of their bounding box. Their position inside the soccer pitch is calculated by projecting this point of the image plane onto a 2D template of the soccer pitch using the field homography matrix. This matrix is previously obtained using the camera calibration model proposed by \cite{Cioppa2021CameraCalibration,Sha2020CameraCalibration}.

\subsection{Unsupervised Player Classification}
\label{sec:playerClassification}

We propose an automatic player classification method that exploits the different field views from a broadcast match to find samples of the five considered categories: \emph{player from team 1} or \emph{team 2}, \emph{goalkeeper A} or \emph{B}, and \emph{referee}. Our proposed method is composed of four main steps that use the information coming from the semantic segmentation and the camera calibration. First, we filter the frames that have a low camera calibration confidence or that show close-up of people. Additionally, we discard detected people with a low detection score and that are visually too small for clustering/classification. Second, we determine which frames correspond to the midfield and cluster the people in them into three groups: \emph{referee}, \emph{player team 1}, \emph{player team 2}. Third, we determine the frames belonging to each goal and consider the person closest to it as being either \emph{goalkeeper A} or \emph{goalkeeper B}. Fourth, we train a convolutional neural network (CNN) using the formed clusters from the two previous steps. This CNN takes as input a masked person and outputs its corresponding category from the five classes. All these steps are fully detailed in the following paragraphs.

\begin{figure}[!t]
\input{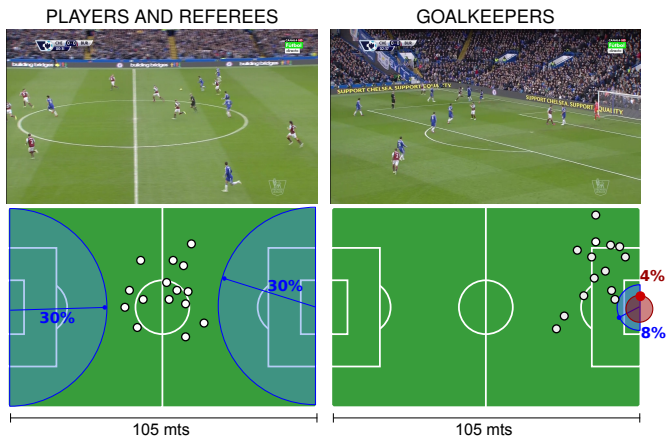}
\caption[]{Examples of selection of \emph{players}, \emph{referees}, and \emph{goalkeepers} using their location in the field. The first column shows a midfield view with \emph{players from both teams} and a \emph{referee} away from the goals by more than 30\% of the length of the field. The second column shows a \emph{goalkeeper} close to the goal by a distance of 4\% the length of the field, and away from the other players by 8\%.}
\label{fig:extractionDistances}
\end{figure}

\paragraph{Frame and people filtering} We filter out the frames that have a low camera calibration confidence score ($\leq0.85$), since the field homography matrix must be accurate in order to obtain the midfield and goal views. Moreover, we also remove those frames that show a close-up of people that might be players or the audience. We consider that a frame shows a close-up of people when at least the bounding box area of one person in the scene is greater than a maximum area threshold.

In addition, we discard segmented people that have a low detection score as they could have been misclassified with other objects on the scene or be a group of occluded persons. We also remove segmented people with a very small bounding box area because they are not useful for the clustering/classification as their uniform colors are not easily distinguishable.

\paragraph{Referee and player teams clustering} We assume that a person in the center of the soccer pitch or midfield is either a \emph{player from team 1}, \emph{team 2}, or the \emph{referee}. This assumption simplifies the clustering problem since goalkeepers rarely appear in the center of the field. In order to get samples of the three considered classes, we first find the frames from the match that show a midfield view. For each frame, we calculate the position of every person on the field using their bounding box from its semantic segmentation and homography matrix of the field. If all persons appearing in a frame are at least 30\% of the field length away from the center of both goals, this frame is considered to have a view of the midfield. An example of this rule for midfield frames is shown in the first column of Figure \ref{fig:extractionDistances}. Although the midfield view avoids photographers and/or team supporters close to the goal or field corners, it might show cameramen, team staff and players warming up on the sidelines. Therefore, we discard detected people who are close to the sidelines using a distance threshold.

The next step is to calculate a color histogram for each person detected in the midfield view frames, as their uniform is their main discriminating characteristic. Each player is represented using a normalized histogram of 64 bins from an 8$\times$8 grid of the a$^*$b$^*$ channels of the L$^*$a$^*$b$^*$ color space, as it proved to be more robust to illumination changes. This histogram is calculated using the pixels inside the (eroded) mask of the semantic segmentation of a person. The purpose of applying an erosion to the segmentation mask is to remove from the contour of a person background pixels belonging to the field or to other people.

The main problem of clustering the people in the midfield without imposing constraints is that sometimes leads to the minority class (i.e., the \emph{referee}) being absorbed by the predominant classes. In order to correctly cluster the actual three classes, we first cluster all players into $n>3$ \emph{prototypes} of them using $k$-means. The value of $n$ should be large enough to form at least one cluster for the \emph{referee} class but not too large to result in several small clusters with outliers. From these cluster prototypes we combinatorially search for the triplet that best represents the \emph{players} and the \emph{referee} classes. Specifically, for each possible triplet we calculate the area of the triangle formed by their centroids. The length of each side of the triangle is the Bhattacharyya distance between the centroids. We choose the triplet of prototypes with the largest triangular area as the one containing the three different classes. Finally, we make these prototypes grow by modelling each using a GMM of $m$ components and iteratively adding samples from the remaining prototypes. This is accomplished in two steps and by considering that the distance $d$ between a sample and a prototype is equal to the median Bhattacharyya distance between the sample and the $m$ components of the prototype. In the first step, a sample is added to a prototype if the distance $d$ between both is close to zero and much smaller than the distance to the other pair of prototypes $d_{1}$ and $d_{2}$ by a threshold $\lambda$, i.e. $d+\lambda<d_{1}$ and $d+\lambda<d_{2}$. In the second step, the GMM of each prototype is recomputed to consider the new additions. Finally, the enlarged prototypes are used to calculate the class boundary by training a support vector machine. The final clusters are calculated conservatively according to the distance between samples and the class boundaries.

\paragraph{Goalkeepers extraction} We assume that the person closest to the goal is \emph{goalkeeper A} or \emph{goalkeeper B}, depending on which side of the field the goal is located. As in the previous case, we compute the locations of each player in the field for all frames using their bounding boxes from the semantic segmentation and homography matrix of the field. For each goal in the field, we look for frames that show scenes that meet two specific conditions. The first condition is that there must be a person at a distance of no more than 4\% the width of the field from the center of the goal. We consider that this person is a goalkeeper, but to make sure that this is the case, the second condition is that the rest of the people are even further away from the goal, at a least a distance of 8\% the width of the field. To illustrate it, the second column of Figure \ref{fig:extractionDistances} shows a potential goalkeeper that fulfills both conditions. Since the conditions do not guarantee to form exact groups of goalkeepers, each group is refined in two steps. First, we compute the a$^*$b$^*$ color histograms from the masked patches of the goalkeepers, as described in the previous case. Second, we calculate the median color histogram value for each \emph{goalkeeper} group and filter out masked patches that have Bhattacharyya distance larger than 0.4. Since camera calibration is not always accurate, in some games the assistant referee is grouped as a goalkeeper and the subsequent filtering results in empty clusters. In this case, the samples that are considered referees are removed from the goalkeepers groups and the filtering is performed again. A sample is considered to be a referee if its Bhattacharyya distance with respect to the referee cluster is lower than a threshold.

\paragraph{Player classification} We train a CNN using as training examples the masked players and referees from the five extracted clusters described above. Specifically, we train from scratch a large MobileNetV3 \cite{Howard2019MobileNetV3} using Adam optimizer \cite{Kingma2015AdamOptimizer}. The training parameters are a starting learning rate $\gamma=1e^{-3}$, $\beta_{1}=0.9$ and $\beta_{2}=0.999$, and a batch size of 96. The network is trained using a plateau schedule with a patience set to 7 epochs for a maximum of 100 epochs. Once the training is finished, all the people in the match are labeled using the CNN.

\begin{figure*}[!t]
\input{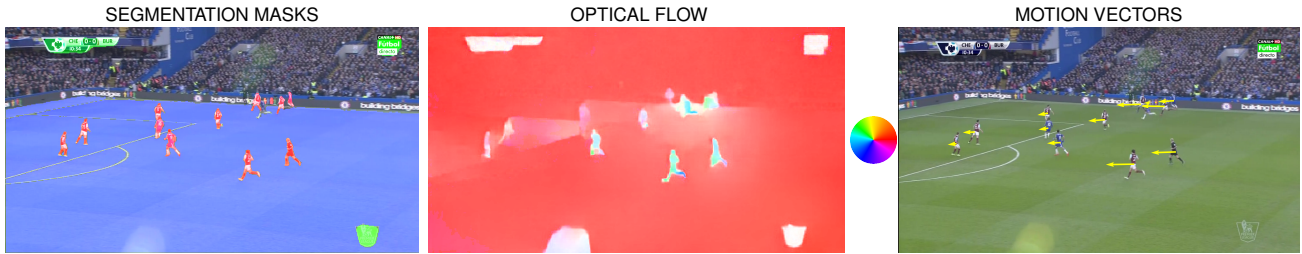}
\caption[]{Example of player motion vector extraction from a frame. The left column shows the calculated segmentation masks for \emph{fixed regions} (green), \emph{field} (blue), and the \emph{players} (red). The middle column shows the obtained optical flow using FlowNet 2.0 CSS \cite{Ilg2017FlowNet2, Reda2017Flownet2Implementation}. The right column shows the extracted motion vectors for each player in yellow.}
\label{fig:motionVectors}
\end{figure*}

\subsection{Player Motion Vector}
\label{sec:playerMotionVector}

Another source of information for our model is the motion vector of each player. This is calculated in four subsequent steps. First, we compute the optical flow for all frames from all matches using FlowNet 2.0 CSS \cite{Ilg2017FlowNet2, Reda2017Flownet2Implementation}. In order to reduce the effect of camera movement and thus the effect of the optical flow from the whole scene over the motion of each player, we calculate the dominant motion vector from the field. This is accomplished by detecting and removing fixed-regions from the match like the scoreboard and segmenting the field in the second and third steps, respectively. In the last step, we calculate the dominant motion vector of each player using the calculated masks from the semantic segmentation. An example of the player motion extraction is shown in Figure \ref{fig:motionVectors}. The next paragraphs describe in detail the last three steps of our method. 

\paragraph{Fixed-regions detection} For each half-match, we obtain the masks of fixed regions by first calculating the mean optical flow from 10\% of the frames uniformly sampled. We expect that the mean optical flow contains fixed and moving regions with values close and different than zero, correspondingly. Then, we perform a $k$-means clustering~\cite{Johnson2022Faiss} by setting $k=4$ and by determining that the masks of the fixed-regions have a centroid with a norm close to zero. Subsequently, for each fixed-region we compute its mean RGB patch using its corresponding mask. The detection of fixed-regions in a frame is a two-step process. For each previously extracted region, \one we obtain a patch of its possible location in the frame and \two calculate the Bhattacharyya distance between this patch and the mean RGB patch.

\paragraph{Field Segmentation} Our field segmentation is based on color quantization of the frame. So first we mask with black color the segmented players and detected fixed-regions. Next, we cluster the color of the pixels using $k$-means~\cite{Johnson2022Faiss} with $k=5$ and consider the largest cluster to be the field. We merge other clusters with the field if the mean squared error between their centroids is close to zero. The merged field pixels form the true values of the field binary mask. A convex hull is obtained from this field mask after performing a binary opening operation and an area size filtering on its regions. The final field mask is calculated by performing a logical \emph{and} operation between the convex hull and the original field mask.

\paragraph{Motion vector} We calculate the motion vector for each segmented player and field using its mask over the optical flow in two steps. In order to avoid cancellations of opposing optical flow vectors, we first compute the magnitude of the dominant optical flow from the second moment matrix as stated in \cite{bigun1991multidimensional,Ballester1998Affine}. Additionally, we calculate its direction by computing the histogram of directions of the four quadrants. In the second step, we subtract the dominant optical flow from the field for each player.

\subsection{Players Graph}
\label{sec:playerGraph}

We encode the players appearing in a frame as nodes in a graph. The features of each node are the five-dimension player classification vector from Section \ref{sec:playerClassification}, its estimated motion vector from Section \ref{sec:playerMotionVector}, the area of its bounding box, its position in the field, the projected center of the frame in the field, and the camera calibration score. In comparison to \cite{Cioppa2021CameraCalibration}, we consider that two players are linked in the graph if there is a distance of 5 meters between them.

Our model processes the graph information in a time window of 15 seconds using 2 \emph{fps}. First, each graph is individually processed using four consecutive dynamic edge graph CNN \cite{Wang2019DynamicGCNN} blocks. Then, the model divides the outputs into two equal parts that serve as input to two independent NetVlad pooling layers ~\cite{Arandjelovic16}. The purpose of this strategy is to preserve the temporal context before and after an action \cite{Giancola2021NetVLADPlusPlus}. The vocabulary size of each NetVlad layer is $k=64$ clusters. Finally, both streams are joined using a fully connected layer that predicts the action.

\section{Experiments}
\label{sec:experiments}

\subsection{Dataset}
\label{sec:dataset}

Our experiments were done using the SoccerNet-V2 dataset~\cite{Deliege2020SoccerNetv2}. Currently, this dataset is the largest of its kind containing 500 soccer broadcast videos, which amounts more than 750 hours. For the task of video summarization, all its videos were annotated using 17 different action categories (for example, \emph{penalty}, \emph{yellow card}, \emph{corner}, etc). Each action annotation is temporally bounded to a single timestamp, in other words, to a specific minute and second in the match. Moreover, these actions are further divided into explicitly \emph{visible} shown or \emph{unshown} events in the video.

Additionally, camera calibration and player detection data at a sampling rate of 2 \emph{fps} were further supplied by \cite{Cioppa2021CameraCalibration}. They trained a student network based on \cite{Sha2020CameraCalibration} to calculate the camera calibration. For the player detection, they provided the bounding boxes extracted using Mask R-CNN~\cite{He2017MaskRCNN}.

\subsection{Player Classification Evaluation}
\label{sec:playerClassificationExperiment}

Since automatic player classification consists of \one clustering players into one of five categories and \two training a classifier using the clustered samples as training/validation splits, the objective of this experiment is to measure the performance of both subtasks. On the one hand, we measure the purity of the samples obtained for each category. On the other hand, we evaluate the classification performance of our method using these splits for training.

\paragraph{Data} We use a single match from the Soccernet-V2 dataset, as it lacks of team player annotations. For the players clustering subtask we use the original provided groundtruth from the single match. Explicitly, following the description of Section \ref{sec:playerClassification}, we cluster the players from the provided 10,800 frames containing 126,287 bounding boxes of detected people. For the player classification subtask we only consider the frames from the second half of the match. Following the filtering process of Section \ref{sec:playerClassification}, we discard frames with a low camera calibration confidence score and frames showing close-ups of people. We manually label the resulting frames into the five considered categories. We did not take into account people not playing in the field, for example, players warming up on the side of the field, coaches, photographers, etc. Additionally, we label players not detected in the groundtruth. We obtain a total of 3,740 annotated frames and 50,256 bounding boxes of the five different classes.

\paragraph{Players clustering evaluation} In order to measure the unsupervised collected data for the training/validation splits, we calculate the purity of the data for each class. In particular, we visually inspect the obtained samples for each class and count the number of samples that correspond to another category or that show more than one person.

\begin{figure}[!t]
\centering
\includegraphics[width=1.0\columnwidth]{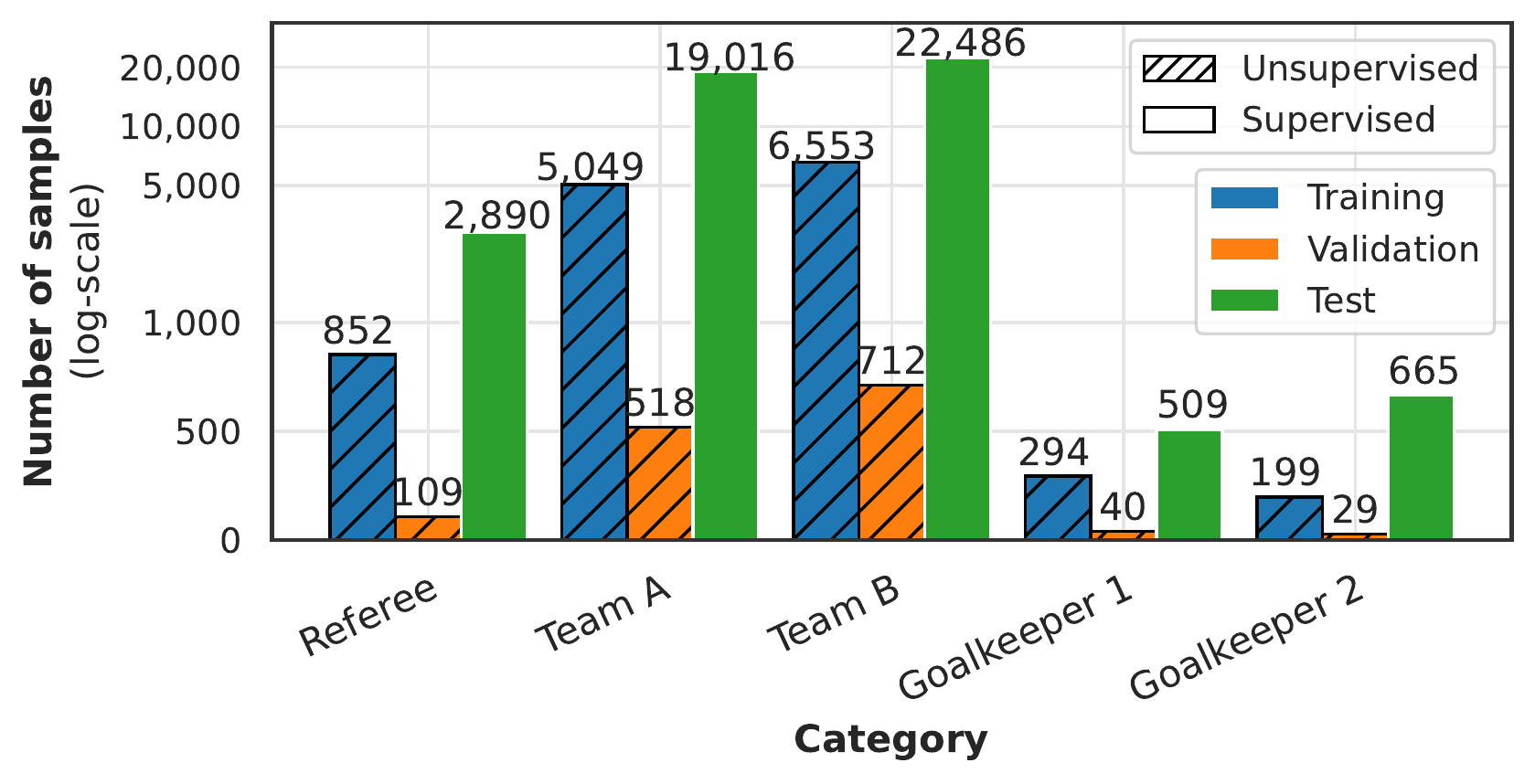}
\caption{The distribution of the data splits for measuring player classification performance. Note that the training and validation splits were obtained in an unsupervised manner by our method.}
\label{fig:trainingValidationSplits}
\end{figure}

\paragraph{Player classification evaluation} We use the annotated bounding boxes from the second half-match as our test split. Since more players were manually added and some people were removed during the annotation process, we only use the bounding boxes with the highest intersection over the union (IoU) coefficient with respect to the groundtruth data. This results in considering 45,566 samples as the test set. The distribution of the resulting data splits for this experiment is shown in Figure \ref{fig:trainingValidationSplits}. To measure the performance we use the classification accuracy metric.

\begin{figure}[!t]
\centering
\input{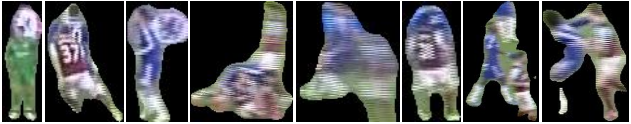}
\caption{Examples of errors in the unsupervised clustered data.}
\label{fig:unsupervisedErrors}
\end{figure}

\paragraph{Results} The performance results for this experiment are shown in Table \ref{tab:playersClassificationPerformance}. For the unsupervised clustering of the training/validation data, the overall purity score is 98.5\%. Some examples of errors in the unsupervised collected data are shown in Figure \ref{fig:unsupervisedErrors}, these examples not only include overlapping players but also other segmented objects near the players. For the classification performance, we obtain an accuracy score of 97.72 for all classes. The results indicate that the natural unbalance of the classes affects the performance of individual classes, specifically for the \emph{goalkeeper} classes. 

\subsection{Node Information and Edges Representation}
\label{sec:ablation}

This experiment serves as an ablation study for the graph representation of the players. The purpose of this experiment is to measure the effects in performance of \one removing the player classification and velocity information from the node, and \two varying the distance between players to consider them linked in the graph.

\begin{table}[!t]
\caption{Performance results for automatic player classification and class purity.}
\label{tab:playersClassificationPerformance}
\centering
\begin{tabular}{lcc}
	\toprule
	\multirow{2}{*}{\textbf{Category}} & \textbf{Class} & \textbf{Classification}\\
	 & \textbf{Purity} & \textbf{Accuracy}\\
	\midrule
	Referee & 95.94 & 97.09\\
	Team A & 99.58 & 98.12\\
	Team B & 97.89 & 97.61\\
	Goalkeeper 1 & 99.40 & 90.57\\
	Goalkeeper 2 & 100.00 & 98.36\\
	\midrule
	\textbf{All} & \textbf{98.50} & \textbf{97.72}\\
	\bottomrule
\end{tabular}

\end{table}

\begin{table}[!t]
\caption{Performance of using different player information on nodes of the graph.}
\label{tab:nodeInformationResults}
\centering
\resizebox{1.00 \columnwidth}{!}{
\begin{tabular}{cccc}
	\toprule
	\B Node & \B \multirow{2}{*}{All} & \B \multirow{2}{*}{visible} & \B \multirow{2}{*}{not-visible}\\
	\B Information & & & \\
	\midrule
	Position & 39.74 & 44.28 & 30.83\\	
	Motion vector+Position & 41.08 & 46.66 & 30.75\\
	Player classification+Position & 41.57 & 45.80 & 30.95\\
	\B All & \B 43.29 & \B 49.16 & \B 31.89 \\ 
	\bottomrule
\end{tabular}
}

\end{table}

\begin{table}[!t]
\caption{Performance of considering distinct distances between players for edge connections.}
\label{tab:distanceResults}
\centering
\begin{tabular}{cccc}
	\toprule
	\B Distance & \B All & \B visible & \B not-visible\\
	\midrule
	\B 5m & \B 43.29 & \B 49.16 & \B 31.89\\
	10m & 43.20 & 49.04 & 31.64\\
	15m & 42.35 & 47.46 & 31.01\\
	20m & 42.17 & 47.90 & 31.22\\
	25m & 43.00 & 47.88 & 31.78\\
	\bottomrule
\end{tabular}

\end{table}

\paragraph{Data} For the ablation study on the node information, we set as baseline information (\textbf{Position}) the location of the players in the soccer pitch, the projected center of the frame in the field, and the camera calibration score. We add to this baseline the calculated motion vector of the player (\textbf{Motion vector+Position}), its predicted team class (\textbf{Player classification+Position}), and both of them (\textbf{All}). In all of these combinations, the considered distance between players was 5 meters.

For the ablation study on the edges, we evaluate a distance between players of $5, 10, \ldots, 25$ meters. These values cover short to long areas of the soccer pitch for the graph, as a radius of 25 meters for a player considers roughly half of the field.

\paragraph{Evaluation} In order to have comparable results with previous works, we use the same public annotated data partitions as presented in \cite{Deliege2020SoccerNetv2}. Moreover, we use the action spotting metric to evaluate the performance as introduced in \cite{Giancola2018SoccerNetv1}. Therefore, we calculate the mean average precision (mAP) across all classes given a temporal Intersection-over-Union (tIoU) of an action. Likewise, this tIoU had a time tolerance $\sigma$ equal to 60 seconds.

\paragraph{Results} The results for using different node information about the players are shown in Table \ref{tab:nodeInformationResults}. They show that the player classification has a slightly better performance for action spotting than motion vector information, and that their added contribution is preserved by reaching an average mAP of 43.29\%. Furthermore, the results of varying the distance between the players of the graph is presented in Table \ref{tab:distanceResults}. It shows that increasing the distance between players reduces the action spotting performance. Thus, we obtain that an adequate distance to link the players in the graph is 5 meters.

\begin{figure}[!t]
\centering
\includegraphics[width=0.8\columnwidth]{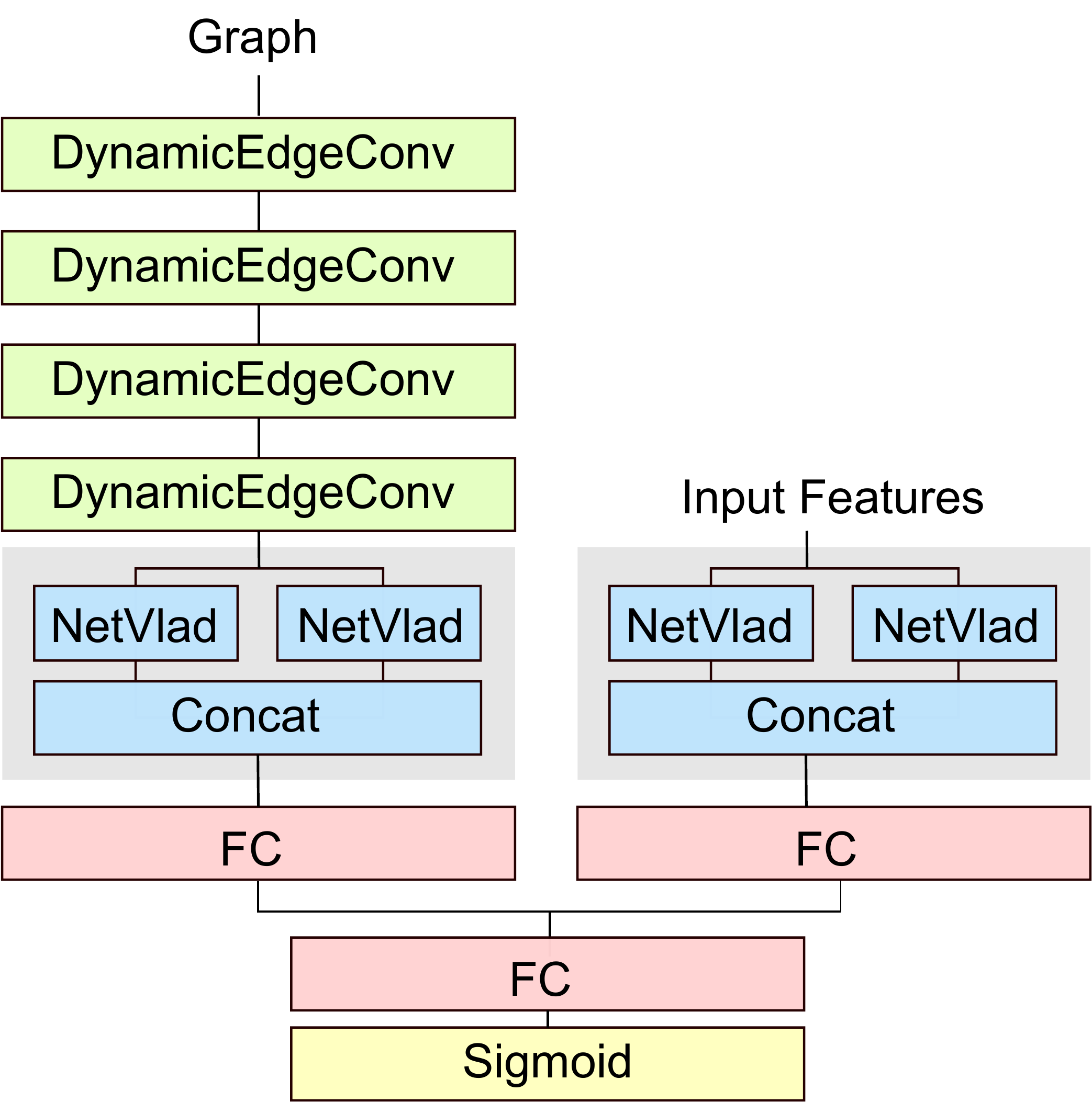}
\caption{A schematic overview of our multimodal architecture that integrates audiovisual streams.}
\label{fig:lateFusionArchitecture}
\end{figure}

\subsection{Different Graph Neural Network Backbones}
\label{sec:gcnBackbones}

The goal of this experiment is to evaluate the performance of using different graph neural network backbones in our architecture. These backbones are put immediately after the graph input and before the pooling mechanism of our architecture, viz. before the NetVlad pooling layers. In particular, we consider substituting our dynamic edge convolutional (\textbf{DynamicEdgeConv}) graph layers \cite{Wang2019DynamicGCNN} by blocks of Graph Convolutional Network (\textbf{GCN}) \cite{Kipf2016GCN} and edge convolutional (\textbf{EdgeConv}) \cite{Wang2019DynamicGCNN} graph layers. Additionally, we also test the graph architecture described by \cite{Cioppa2021CameraCalibration} that uses 14 consecutive deep graph convolutional (\textbf{DeeperGCN}) layers introduced by \cite{Li2021DeepGCNs}.

\paragraph{Data} For all tested architectures we use the same node information described throughout Section \ref{sec:method}, that is we use the location, player classification, and motion vector to represent a player. Moreover, we consider that two players share an edge in the graph if the distance between them is 5 meters.

\paragraph{Evaluation} We use the same data partitions and action spotting metric described in the experiment of Section \ref{sec:ablation}.

\paragraph{Results} The results for this experiment are shown in Table \ref{tab:backboneResults}. They indicate that the best architecture configuration is to use DynamicEdgeConv layers as backbone. This configuration has an action spotting score of 43.29\% and it is followed by DeeperGCN \cite{Li2021DeepGCNs}, the architecture used as backbone in \cite{Cioppa2021CameraCalibration}.

\begin{table}[!t]
\caption{Performance of using different graph neural network backbones.}
\label{tab:backboneResults}
\centering
\resizebox{1.00 \columnwidth}{!}{
\begin{tabular}{lccc}
	\toprule
	\B GCN Backbone & \B All & \B visible & \B not-visible\\
	\midrule
	EdgeConv \cite{Wang2019DynamicGCNN} & 36.99 & 42.09& 29.29 \\
	GCN \cite{Kipf2016GCN} & 40.77 & 45.73 & 30.58\\
	DeeperGCN \cite{Li2021DeepGCNs,Cioppa2021CameraCalibration} & 42.43 & 47.37 & \B 33.21\\
	\B DynamicEdgeConv \cite{Wang2019DynamicGCNN} & \B 43.29 & \B 49.16 & 31.89\\
	\bottomrule
\end{tabular}
}
\end{table}

\begin{table}[!t]
\caption{Performance of combining different modalities to our graph-based model.}
\label{tab:multimodalResults}
\centering
\resizebox{1.00 \columnwidth}{!}{
\begin{tabular}{lccc}
	\toprule
	\B Modalities & \B All & \B visible & \B not-visible\\
	\midrule
	Audio & 31.59 & 37.33 & 19.44\\
	Graph &43.29 &49.16 & 31.89\\
	RGB&51.28 &56.17 & 34.62\\
	Audio+Graph&49.23 &53.45 & 38.14\\
	RGB+Graph&51.53 &56.47 & 35.65\\
	RGB+Audio&53.55 &61.32 & 35.08\\
	\B RGB+Audio+Graph& \B57.83 &\B63.92 & \B42.78\\
	\bottomrule
\end{tabular}
}
\end{table}

\begin{table*}[ht]
\scriptsize
    \caption{\Sota comparison. We report the results for action spotting (Average-mAP \%) on \SN2~\cite{Deliege2020SoccerNetv2}. Best results are indicated in {\bf bold} and second best are \underline{underlined}.}
    \centering
    \setlength\arrayrulewidth{0.005em}
    \setlength{\tabcolsep}{1.2pt}
    \renewcommand{\arraystretch}{1.5}
    \resizebox{\linewidth}{!}{
    \begin{tabular}{l!{\boldVerticalLine}c | c|c!{\boldVerticalLine}c|c|c|c|c|c|c|c|c|c|c|c|c|c|c|c|c}
    \B\B\fontsize{7}{10}\selectfont Method &  \begin{turn}{90}\B\B\fontsize{7}{10}\selectfont All\end{turn} &  \begin{turn}{90}\B\B\fontsize{7}{10}\selectfont visible\end{turn}   &  \begin{turn}{90}\B\B\fontsize{7}{10}\selectfont not visible\end{turn}  & \begin{turn}{90}\B\B\fontsize{7}{10}\selectfont Ball out \end{turn} & \begin{turn}{90}\B\B\fontsize{7}{10}\selectfont Throw-in\end{turn} & \begin{turn}{90}\B\B\fontsize{7}{10}\selectfont Foul \end{turn} & \begin{turn}{90}\B\B\fontsize{7}{10}\selectfont Ind. free-kick \end{turn} & \begin{turn}{90}\B\B\fontsize{7}{10}\selectfont Clearance \end{turn} & \begin{turn}{90}\B\B\fontsize{7}{10}\selectfont Shots on tar. \end{turn} & \begin{turn}{90}\B\B\fontsize{7}{10}\selectfont Shots off tar. \end{turn} & \begin{turn}{90}\B\B\fontsize{7}{10}\selectfont Corner \end{turn} & \begin{turn}{90}\B\B\fontsize{7}{10}\selectfont Substitution \end{turn} & \begin{turn}{90}\B\B\fontsize{7}{10}\selectfont Kick-off \end{turn} & \begin{turn}{90}\B\B\fontsize{7}{10}\selectfont Yellow card \end{turn} & \begin{turn}{90}\B\B\fontsize{7}{10}\selectfont Offside \end{turn} & \begin{turn}{90}\B\B\fontsize{7}{10}\selectfont Dir. free-kick \end{turn} & \begin{turn}{90}\B\B\fontsize{7}{10}\selectfont Goal \end{turn} & \begin{turn}{90}\B\B\fontsize{7}{10}\selectfont Penalty \end{turn} & \begin{turn}{90}\B\B\fontsize{7}{10}\selectfont Yel.$\to$Red \end{turn} & \begin{turn}{90}\B\B\fontsize{7}{10}\selectfont Red card\end{turn} \\
    \specialrule{.1em}{.1em}{.1em}
AudioVid~\cite{Vanderplaetse2020Improved}   &  39.9 &  43.0 &  23.3 &  54.3 &  50.0 &  55.5 &  22.7 &  46.7 &  26.5 &  21.4 &  66.0 &  54.0 &  52.9 &  35.2 &  24.3 &  46.7 &  69.7 &  52.1 &  0.0 &  0.0 \\ \hline
CameraCalibration~\cite{Cioppa2021CameraCalibration}&46.80 &- & -  & 60.40 & 34.80 & 50.40 & 33.60 & 78.70 & 32.50 & 33.00 & 76.20 & 45.50 & 59.90 & 50.50 & 55.40 & 48.60 & 56.20 & 68.70 & 8.50 & 3.10 \\ \hline
NetVLAD++~\cite{Giancola2021NetVLADPlusPlus}&53.40 &59.41 & 34.97  & 68.69 & 62.15 & 56.66 & 39.22 & 79.74 & 39.22 & 40.98 & 71.58 & 44.36 & 69.01 & 79.28 & 57.05 & 57.77 & 64.20 & 70.30 & 3.96 & 3.69 \\ \hline
RGB+Audio+Graph (Ours)&57.83 &63.92 & \underline{42.78}  & \underline{76.06} & \underline{67.19} & 73.65 & 48.15 & 63.43 & \underline{50.01} & 45.03 & \underline{85.14} & \textbf{71.29} & 66.60 & 50.29 & 34.93 & 62.45 & \underline{83.21} & \underline{85.34} & 13.91 & 6.40 \\ \hline
AImageLab-RMS~\cite{Tomei2021Rmsnet}& \underline{63.49} & \underline{68.88} & 38.02  & 71.28 & 65.19 & \underline{76.76} & \underline{58.32} & \underline{88.80} & 37.94 & \underline{50.12} & 77.60 & 64.29 & \underline{67.99} & \underline{89.20} & \underline{76.24} & \underline{66.83} & 69.27 & 58.67 & \textbf{30.00} & \underline{30.83} \\ \hline
Vidpress Sports~\cite{Zhou2021FeatureCombination}&\textbf{73.77} &\textbf{79.28} & \textbf{47.84}  & \textbf{78.98} & \textbf{82.83} & \textbf{79.11} & \textbf{67.97} & \textbf{88.92} & \textbf{66.28} & \textbf{65.69} & \textbf{86.81} & \underline{71.22} & \textbf{87.81} & \textbf{93.63} & \textbf{76.50} & \textbf{75.05} & \textbf{84.23} & \textbf{87.87} & \underline{25.62} & \textbf{35.60} \\
\specialrule{.1em}{.1em}{.1em}
\end{tabular}
}
\label{tab:results}
\end{table*}

\subsection{Comparison with the State of the Art}
\label{sec:comparison}

In this experiment, we compare our model with state-of-the-art methods on the same dataset. Since our model only considers local scene information, we also test combining it with global information features. These features are combined with our architecture using a late fusion approach \cite{Karpathy2014LargeScale}, as shown in Figure \ref{fig:lateFusionArchitecture}. In this architecture all the weights were frozen except the last fully-connected (FC) layer. We consider features extracted from the audio and frames of the video sequence as additional modalities of our graph-based model. Moreover, we measure the performance of all the input combinations of these audiovisual modalities.

\paragraph{Data} For our graph-based stream (\textbf{Graph}), we use the all the players information detailed in Section \ref{sec:method}. In the case of the audio stream (\textbf{Audio}), the short-time Fourier transform (STFT) spectrogram is computed from the resampled audio signal at 16 KHz with a window size of 25 ms, a window hop of 10 ms, and a periodic Hann window. An stabilized log-mel spectrogram is calculated by mapping the spectrogram to 64 mel bins covering the range 125-7500 Hz. This final spectrogram is used as input for the VGGish network ~\cite{Gemmeke2017AudioSet}, a CNN pretrained on AudioSet. An audio feature vector covers a non-overlapping half-second chunk and corresponds to the vector of dimension 512 of the last fully connected (FC) layer. These features were originally extracted from SoccerNet-V2 dataset by \cite{Vanderplaetse2020Improved}. For the visual stream (\textbf{RGB}), we use the features provided by \cite{Deliege2020SoccerNetv2}. The features are obtained by first using a pretrained ResNet-152~\cite{He2016ResNet} network and, subsequently, applying a principal component analysis (PCA) on its FC layer. This reduces its dimension from 2,048 to 512. The sampling frequency for all modalities is 2 frames per second.

\paragraph{Evaluation} In order to compare our method with other published results, we use the same public splits and action spotting metric described in the experiment of Section \ref{sec:ablation}.

\paragraph{Results} The results of using different input modalities are shown in Table \ref{tab:multimodalResults}. They show that our graph model is better than the standalone audio stream but not than the RGB stream. This may be because broadcast videos contain interleave shots of field-views, replays, and close-ups of the players and the audience, and a graph is only created from field-views. Furthermore, the performance increases when the three streams (audio, RGB and graph) are combined. Finally, the results of our comparison with the state-of-the-art methods are shown in Table \ref{tab:results}. We obtain an overall action spotting performance of $57.83$\% average-mAP. This result indicates that our method outperforms a similar graph-based method for action spotting, namely the CameraCalibration~\cite{Cioppa2021CameraCalibration}. Additionally, our method has comparable results to the state-of-the-art methods that are more computing intensive as Vidpress Sports~\cite{Zhou2021FeatureCombination}, as it is in the second (or first) position for seven categories out of seventeen (e.g. \emph{substitution} or \emph{ball-out}).

\section{Conclusions}
\label{sec:conclusions}

We present a method for soccer action spotting based on a graph representation of the players in the field. In this graph, a player is represented as a node that encodes its location, motion vector, and player classification. The players share an edge in the graph according to the physical distance between them. All the node information is extracted from the segmentation of the players and their location given by camera calibration information. In particular, we propose an automatic player classification method that considers five different classes: \emph{player from team 1} or \emph{team 2}, \emph{goalkeeper A} or \emph{B}, and \emph{referee}. This method relies on color histograms from the segmented players and their position in the field. We extensively evaluate our methods in the SoccerNet-v2 dataset, the largest collection of broadcast soccer videos. For our player classification method we obtain an accuracy of $97.72$\% in our annotated benchmark. Furthermore, we obtain an overall action spotting performance of $57.83$\% average-mAP by combining our method with audiovisual modalities. Our method outperforms similar graph-based approaches and has comparable results with state-of-the-art methods.

\paragraph{Acknowledgments} The authors acknowledge support by MICINN/FEDER UE project, ref. PID2021-127643NB-I00, H2020-MSCA-RISE-2017 project, ref. 777826 NoMADS, and ReAViPeRo network, ref. RED2018-102511-T.

{\small
\bibliographystyle{ieee_fullname}

\begin{thebibliography}{10}\itemsep=-1pt

\bibitem{Anzer2022TacticalPatterns}
Gabriel Anzer, Pascal Bauer, Ulf Brefeld, and Dennis Faßmeyer.
\newblock Detection of tactical patterns using semi-supervised graph neural
  networks.
\newblock 03 2022.

\bibitem{Arandjelovic16}
R. Arandjelovi\'c, P. Gronat, A. Torii, T. Pajdla, and J. Sivic.
\newblock {NetVLAD}: {CNN} architecture for weakly supervised place
  recognition.
\newblock In {\em IEEE Conference on Computer Vision and Pattern Recognition},
  2016.

\bibitem{Arbues2020Always}
A. Arbués-Sangüesa, A. Martín, J. Fernández, C. Rodríguez, G. Haro, and C.
  Ballester.
\newblock Always look on the bright side of the field: Merging pose and
  contextual data to estimate orientation of soccer players.
\newblock In {\em 2020 IEEE International Conference on Image Processing
  (ICIP)}, pages 1506--1510, 2020.

\bibitem{Ballester1998Affine}
Coloma Ballester and Manuel Gonz\'{a}lez.
\newblock Affine invariant texture segmentation and shape from texture by
  variational methods.
\newblock {\em J. Math. Imaging Vis.}, 9(2):141–171, sep 1998.

\bibitem{Bialkowski2013TeamActivities}
Alina Bialkowski, Patrick Lucey, Peter Carr, Simon Denman, Iain Matthews, and
  Sridha Sridharan.
\newblock Recognising team activities from noisy data.
\newblock In {\em 2013 IEEE Conference on Computer Vision and Pattern
  Recognition Workshops}, pages 984--990, 2013.

\bibitem{bigun1991multidimensional}
Josef Big{\"u}n, Goesta~H. Granlund, and Johan Wiklund.
\newblock Multidimensional orientation estimation with applications to texture
  analysis and optical flow.
\newblock {\em IEEE Transactions on Pattern Analysis \& Machine Intelligence},
  13(08):775--790, 1991.

\bibitem{Baldu2019Using}
Javier~M. Buldú, Javier Busquets, Johann~H. Martínez, José~L.
  Herrera-Diestra, Ignacio Echegoyen, Javier Galeano, and Jordi Luque.
\newblock Using network science to analyse football passing networks: Dynamics,
  space, time, and the multilayer nature of the game.
\newblock {\em Frontiers in Psychology}, 9, 2018.

\bibitem{Chen2018CameraSelection}
Jianhui Chen, Lili Meng, and James~J. Little.
\newblock Camera selection for broadcasting soccer games.
\newblock In {\em 2018 IEEE Winter Conference on Applications of Computer
  Vision (WACV)}, pages 427--435, 2018.

\bibitem{Cioppa2020context}
Anthony Cioppa, Adrien Deli{\`e}ge, Silvio Giancola, Bernard Ghanem, Marc~Van
  Droogenbroeck, Rikke Gade, and Thomas~B. Moeslund.
\newblock A context-aware loss function for action spotting in soccer videos.
\newblock In {\em IEEE Conference on Computer Vision and Pattern Recognition
  (CVPR)}, pages 13126--13136, 2020.

\bibitem{Cioppa2021CameraCalibration}
Anthony Cioppa, Adrien Deli{\`e}ge, Floriane Magera, Silvio Giancola, Olivier
  Barnich, Bernard Ghanem, and {Marc Van} Droogenbroeck.
\newblock Camera calibration and player localization in soccernet-v2 and
  investigation of their representations for action spotting.
\newblock In {\em 2021 IEEE/CVF Conference on Computer Vision and Pattern
  Recognition Workshops (CVPRW)}. IEEE, 2021.

\bibitem{Coimbra2018Shape}
Danilo~B. Coimbra, Tácito Trindade~de Araújo Tiburtino~Neves, Alexandru~C.
  Telea, and Fernando~V. Paulovich.
\newblock The shape of the game.
\newblock In {\em 2018 31st SIBGRAPI Conference on Graphics, Patterns and
  Images (SIBGRAPI)}, pages 134--141, 2018.

\bibitem{FIFA2018AudienceSummary}
Fédération~Internationale de Football~Association.
\newblock 2018 fifa world cup russia: Global broadcast and audience summary,
  Dec. 2018.

\bibitem{Deliege2020SoccerNetv2}
Adrien Deliège, Anthony Cioppa, Silvio Giancola, Meisam~J. Seikavandi,
  Jacob~V. Dueholm, Kamal Nasrollahi, Bernard Ghanem, Thomas~B. Moeslund, and
  Marc~Van Droogenbroeck.
\newblock Soccernet-v2 : A dataset and benchmarks for holistic understanding of
  broadcast soccer videos.
\newblock In {\em The IEEE Conference on Computer Vision and Pattern
  Recognition (CVPR) Workshops}, June 2021.

\bibitem{FPFD2019FIFAReport}
FIFA Professional~Football Department and International~Centre for
  Sports~Studies.
\newblock Fifa: Professional football report 2019, Dec. 2019.

\bibitem{DOrazio2009BSAS}
Tiziana D'Orazio, Marco Leo, Paolo Spagnolo, Pier~Luigi Mazzeo, Nicola Mosca,
  Massimiliano Nitti, and Arcangelo Distante.
\newblock An investigation into the feasibility of real-time soccer offside
  detection from a multiple camera system.
\newblock {\em IEEE Transactions on Circuits and Systems for Video Technology},
  19(12):1804--1818, 2009.

\bibitem{Gemmeke2017AudioSet}
Jort~F. Gemmeke, Daniel P.~W. Ellis, Dylan Freedman, Aren Jansen, Wade
  Lawrence, R.~Channing Moore, Manoj Plakal, and Marvin Ritter.
\newblock Audio set: An ontology and human-labeled dataset for audio events.
\newblock In {\em 2017 IEEE International Conference on Acoustics, Speech and
  Signal Processing (ICASSP)}, pages 776--780, 2017.

\bibitem{Giancola2018SoccerNetv1}
Silvio Giancola, Mohieddine Amine, Tarek Dghaily, and Bernard Ghanem.
\newblock {SoccerNet: A Scalable Dataset for Action Spotting in Soccer Videos}.
\newblock In {\em IEEE Conference on Computer Vision and Pattern Recognition
  (CVPR) Workshops}, pages 1711--1721, June 2018.

\bibitem{Giancola2021NetVLADPlusPlus}
Silvio Giancola and Bernard Ghanem.
\newblock Temporally-aware feature pooling for action spotting in video
  broadcasts.
\newblock In {\em The IEEE Conference on Computer Vision and Pattern
  Recognition (CVPR) Workshops}, June 2021.

\bibitem{He2017MaskRCNN}
Kaiming He, Georgia Gkioxari, Piotr Dollár, and Ross Girshick.
\newblock Mask r-cnn.
\newblock In {\em 2017 IEEE International Conference on Computer Vision
  (ICCV)}, pages 2980--2988, 2017.

\bibitem{He2016ResNet}
Kaiming He, Xiangyu Zhang, Shaoqing Ren, and Jian Sun.
\newblock Deep residual learning for image recognition.
\newblock In {\em 2016 IEEE Conference on Computer Vision and Pattern
  Recognition (CVPR)}, pages 770--778, 2016.

\bibitem{Howard2019MobileNetV3}
Andrew Howard, Mark Sandler, Grace Chu, Liang-Chieh Chen, Bo Chen, Mingxing
  Tan, Weijun Wang, Yukun Zhu, Ruoming Pang, Vijay Vasudevan, Quoc~V. Le, and
  Hartwig Adam.
\newblock Searching for mobilenetv3.
\newblock In {\em Proceedings of the IEEE/CVF International Conference on
  Computer Vision (ICCV)}, October 2019.

\bibitem{Ilg2017FlowNet2}
E. Ilg, N. Mayer, T. Saikia, M. Keuper, A. Dosovitskiy, and T. Brox.
\newblock Flownet 2.0: Evolution of optical flow estimation with deep networks.
\newblock In {\em IEEE Conference on Computer Vision and Pattern Recognition
  (CVPR)}, Jul 2017.

\bibitem{Istasse2019Associative}
Maxime Istasse, Julien Moreau, and Christophe~De Vleeschouwer.
\newblock Associative embedding for team discrimination.
\newblock In {\em {IEEE} Conference on Computer Vision and Pattern Recognition
  Workshops, {CVPR} Workshops 2019, Long Beach, CA, USA, June 16-20, 2019},
  pages 2477--2486. Computer Vision Foundation / {IEEE}, 2019.

\bibitem{Johnson2022Faiss}
Jeff Johnson, Matthijs Douze, and Hervé Jégou.
\newblock Billion-scale similarity search with gpus.
\newblock {\em IEEE Transactions on Big Data}, 7(3):535--547, 2021.

\bibitem{Karpathy2014LargeScale}
Andrej Karpathy, George Toderici, Sanketh Shetty, Thomas Leung, Rahul
  Sukthankar, and Li Fei-Fei.
\newblock Large-scale video classification with convolutional neural networks.
\newblock In {\em IEEE Conference on Computer Vision and Pattern Recognition
  (CVPR)}, 2014.

\bibitem{Kingma2015AdamOptimizer}
Diederik~P. Kingma and Jimmy Ba.
\newblock Adam: {A} method for stochastic optimization.
\newblock In Yoshua Bengio and Yann LeCun, editors, {\em 3rd International
  Conference on Learning Representations, {ICLR} 2015, San Diego, CA, USA, May
  7-9, 2015, Conference Track Proceedings}, 2015.

\bibitem{Kipf2016GCN}
Thomas~N. Kipf and Max Welling.
\newblock Semi-supervised classification with graph convolutional networks.
\newblock In {\em 5th International Conference on Learning Representations,
  {ICLR} 2017, Toulon, France, April 24-26, 2017, Conference Track
  Proceedings}. OpenReview.net, 2017.

\bibitem{Kirillov2020PointRend}
Alexander Kirillov, Yuxin Wu, Kaiming He, and Ross~B. Girshick.
\newblock Pointrend: Image segmentation as rendering.
\newblock In {\em 2020 {IEEE/CVF} Conference on Computer Vision and Pattern
  Recognition, {CVPR} 2020, Seattle, WA, USA, June 13-19, 2020}, pages
  9796--9805. Computer Vision Foundation / {IEEE}, 2020.

\bibitem{Koshkina2021contrastive}
Maria Koshkina, Hemanth Pidaparthy, and James~H Elder.
\newblock Contrastive learning for sports video: Unsupervised player
  classification.
\newblock In {\em Proceedings of the IEEE/CVF Conference on Computer Vision and
  Pattern Recognition}, pages 4528--4536, 2021.

\bibitem{KPMG2020EuropeanChampions}
KPMG.
\newblock The european champions report 2020, Jan. 2020.

\bibitem{Li2021DeepGCNs}
Guohao Li, Matthias M{\"u}ller, Guocheng Qian, Itzel Carolina~Delgadillo Perez,
  Abdulellah Abualshour, Ali~Kassem Thabet, and Bernard Ghanem.
\newblock Deepgcns: Making gcns go as deep as cnns.
\newblock {\em IEEE Transactions on Pattern Analysis and Machine Intelligence},
  2021.

\bibitem{Manafifard2016MultiplayerDetection}
M. Manafifard, H. Ebadi, and H.A. Moghaddam.
\newblock Multi-player detection in soccer broadcast videos using a blob-guided
  particle swarm optimization method.
\newblock {\em Multimedia Tools Appl.}, pages 1--30, 2016.
\newblock cited By 1.

\bibitem{Mazzeo2012football}
Pier~Luigi Mazzeo, Paolo Spagnolo, Marco Leo, and Tiziana D'Orazio.
\newblock Football players classification in a multi-camera environment.
\newblock In Jacques Blanc-Talon, Don Bone, Wilfried Philips, Dan Popescu, and
  Paul Scheunders, editors, {\em Advanced Concepts for Intelligent Vision
  Systems}, pages 143--154, Berlin, Heidelberg, 2010. Springer Berlin
  Heidelberg.

\bibitem{Montanes2012RealTime}
Miguel~Angel Monta{\~{n}}{\'e}s~Laborda, Enrique~F. Torres~Moreno, Jes{\'u}s
  Mart{\'i}nez~del Rinc{\'o}n, and Jos{\'e}~El{\'i}as Herrero~Jaraba.
\newblock Real-time gpu color-based segmentation of football players.
\newblock {\em Journal of Real-Time Image Processing}, 7(4):267--279, Dec 2012.

\bibitem{Pappalardo2019PublicData}
Luca Pappalardo, Paolo Cintia, Alessio Rossi, Emanuele Massucco, Paolo
  Ferragina, Dino Pedreschi, and Fosca Giannotti.
\newblock A public data set of spatio-temporal match events in soccer
  competitions.
\newblock {\em Scientific Data}, 6(1):236, Oct 2019.

\bibitem{Passos2010Networks}
P Passos, K Davids, D Ara{\'u}jo, N Paz, J Mingu{\'e}ns, and J Mendes.
\newblock Networks as a novel tool for studying team ball sports as complex
  social systems.
\newblock {\em J Sci Med Sport}, 14(2):170--176, Dec. 2010.

\bibitem{Qi2020stagNet}
Mengshi Qi, Yunhong Wang, Jie Qin, Annan Li, Jiebo Luo, and Luc Van~Gool.
\newblock stagnet: An attentive semantic rnn for group activity and individual
  action recognition.
\newblock {\em IEEE Transactions on Circuits and Systems for Video Technology},
  30(2):549--565, 2020.

\bibitem{Reda2017Flownet2Implementation}
Fitsum Reda, Robert Pottorff, Jon Barker, and Bryan Catanzaro.
\newblock flownet2-pytorch: Pytorch implementation of flownet 2.0: Evolution of
  optical flow estimation with deep networks.
\newblock https://github.com/NVIDIA/flownet2-pytorch, 2017.

\bibitem{Sanabria2020Profiling}
Melissa Sanabria, Frédéric Precioso, and Thomas Menguy.
\newblock Profiling actions for sport video summarization: An attention signal
  analysis.
\newblock In {\em 2020 IEEE 22nd International Workshop on Multimedia Signal
  Processing (MMSP)}, pages 1--6, 2020.

\bibitem{Sanabria2019Deep}
Melissa Sanabria, Sherly, Fr\'{e}d\'{e}ric Precioso, and Thomas Menguy.
\newblock A deep architecture for multimodal summarization of soccer games.
\newblock In {\em Proceedings Proceedings of the 2nd International Workshop on
  Multimedia Content Analysis in Sports}, MMSports '19, page 16–24, New York,
  NY, USA, 2019. Association for Computing Machinery.

\bibitem{Sha2020CameraCalibration}
Long Sha, Jennifer Hobbs, Panna Felsen, Xinyu Wei, Patrick Lucey, and Sujoy
  Ganguly.
\newblock End-to-end camera calibration for broadcast videos.
\newblock In {\em Proceedings of the IEEE/CVF Conference on Computer Vision and
  Pattern Recognition (CVPR)}, June 2020.

\bibitem{Stockl2021Offensive}
Michael Stöckl, Thomas Seidl, Daniel Marley, and Paul Power.
\newblock Making offensive play predictable -using a graph convolutional
  network to understand defensive performance in soccer.
\newblock 04 2021.

\bibitem{Tomei2021Rmsnet}
Matteo Tomei, Lorenzo Baraldi, Simone Calderara, Simone Bronzin, and Rita
  Cucchiara.
\newblock Rms-net: Regression and masking for soccer event spotting, 2021.

\bibitem{Tong2011LabelingTracking}
Xiaofeng Tong, Jia Liu, Tao Wang, and Yimin Zhang.
\newblock Automatic player labeling, tracking and field registration and
  trajectory mapping in broadcast soccer video.
\newblock {\em ACM Trans. Intell. Syst. Technol.}, 2(2), feb 2011.

\bibitem{Tran2012LongView}
Quang Tran, An Tran, Tien~Ba Dinh, and Duc Duong.
\newblock Long-view player detection framework algorithm in broadcast soccer
  videos.
\newblock In De-Shuang Huang, Yong Gan, Phalguni Gupta, and M.~Michael Gromiha,
  editors, {\em Advanced Intelligent Computing Theories and Applications. With
  Aspects of Artificial Intelligence}, pages 557--564, Berlin, Heidelberg,
  2012. Springer Berlin Heidelberg.

\bibitem{Vanderplaetse2020Improved}
Bastien Vanderplaetse and Stephane Dupont.
\newblock Improved soccer action spotting using both audio and video streams.
\newblock In {\em IEEE Conference on Computer Vision and Pattern Recognition
  (CVPR) Workshops}, pages 3921--3931, June 2020.

\bibitem{Wang2019DynamicGCNN}
Yue Wang, Yongbin Sun, Ziwei Liu, Sanjay~E. Sarma, Michael~M. Bronstein, and
  Justin~M. Solomon.
\newblock Dynamic graph cnn for learning on point clouds.
\newblock 38(5), oct 2019.

\bibitem{Zhou2021FeatureCombination}
Xin Zhou, Le Kang, Zhiyu Cheng, Bo He, and Jingyu Xin.
\newblock Feature combination meets attention: Baidu soccer embeddings and
  transformer based temporal detection, 2021.

\end{thebibliography}

}

\end{document}